%% file: main.tex
\documentclass[10pt,twocolumn,letterpaper]{article}

\usepackage[pagenumbers]{iccv} 


\definecolor{iccvblue}{rgb}{0.21,0.49,0.74}
\usepackage[pagebackref,breaklinks,colorlinks,allcolors=iccvblue]{hyperref}
\usepackage{url}            
\usepackage{booktabs}       
\usepackage{amsfonts}       
\usepackage{nicefrac}       
\usepackage{xcolor}         
\usepackage{amsthm}
\usepackage{graphicx}
\usepackage{algorithm}
\usepackage{algorithmic}
\usepackage{multirow}
\usepackage{fbox}
\usepackage{indentfirst}
\usepackage{bm}
\usepackage{colortbl}
\usepackage{makecell}

\def\paperID{778} 
\def\confName{ICCV}
\def\confYear{2025}

\definecolor{blue_1}{RGB}{0,112,192}

\title{SMoLoRA: Exploring and Defying Dual Catastrophic Forgetting in Continual Visual Instruction Tuning}

\author{Ziqi Wang\textsuperscript{1}, Chang Che\textsuperscript{1}, Qi Wang\textsuperscript{2}, Yangyang Li\textsuperscript{3}, Zenglin Shi\textsuperscript{1}\thanks{Corresponding author: zenglin.shi@hfut.edu.cn} , Meng Wang\textsuperscript{1}\\
{\textsuperscript{1}Hefei University of Technology }
{\textsuperscript{2}Tsinghua University }
{\textsuperscript{3}Academy of Cyber}
}

\begin{document}
\maketitle
\input{sec_our/0_abstract}    
\input{sec_our/1_intro}

\input{sec_our/2_related}

\input{sec_our/3_method}

\input{sec_our/4_benchmark}

\input{sec_our/5_exp}

\input{sec_our/6_conclusion}

{
    \small
    \bibliographystyle{ieeenat_fullname}
    \bibliography{main}
}
\input{sec_our/X_suppl}

\end{document}

%% file: sec_our/0_abstract.tex
\begin{abstract}
Visual instruction tuning (VIT) enables multimodal large language models (MLLMs) to effectively handle a wide range of vision tasks by framing them as language-based instructions. Building on this, continual visual instruction tuning (CVIT) extends the capability of MLLMs to incrementally learn new tasks, accommodating evolving functionalities. While prior work has advanced CVIT through the development of new benchmarks and approaches to mitigate catastrophic forgetting, these efforts largely follow traditional continual learning paradigms, neglecting the unique challenges specific to CVIT. We identify a dual form of catastrophic forgetting in CVIT, where MLLMs not only forget previously learned visual understanding but also experience a decline in instruction following abilities as they acquire new tasks. To address this, we introduce the Separable Mixture of Low-Rank Adaptation (SMoLoRA) framework, which employs separable routing through two distinct modules—one for visual understanding and another for instruction following. This dual-routing design enables specialized adaptation in both domains, preventing forgetting while improving performance. Furthermore, we propose a new CVIT benchmark that goes beyond existing benchmarks by additionally evaluating a model's ability to generalize to unseen tasks and handle diverse instructions across various tasks. Extensive experiments demonstrate that SMoLoRA outperforms existing methods in mitigating dual forgetting, improving generalization to unseen tasks, and ensuring robustness in following diverse instructions. Code is available at https://github.com/Minato-Zackie/SMoLoRA.

\end{abstract}

%% file: sec_our/1_intro.tex
\section{Introduction}
\label{sec:intro}

Building on the foundation of Large Language Models (LLMs) \cite{touvron2023llama, kaddour2023challenges}, Multimodal Large Language Models (MLLMs), \eg, \cite{dai2023instructblip, liu2024visual, zhu2023minigpt}, have demonstrated strong performance across a range of tasks, such as visual question answering (VQA) \cite{chen2024mllm, lee2024visual}, image captioning \cite{awadalla2023openflamingo, liu2024visual}, and visual reasoning \cite{huang2023language, wang2024exploring}. MLLMs are typically trained through a multi-stage process \cite{zhu2023minigpt, liu2024visual}. In the pre-training stage, the model acquires general knowledge across multiple modalities, while the visual instruction tuning stage enables it to learn and address various vision-related tasks by framing these tasks as language instructions. This approach transforms the model into a versatile, general-purpose multimodal system capable of following a wide range of user-defined instructions to solve specific tasks.

\begin{figure}[t]
  \centering
   \includegraphics[width=\linewidth]{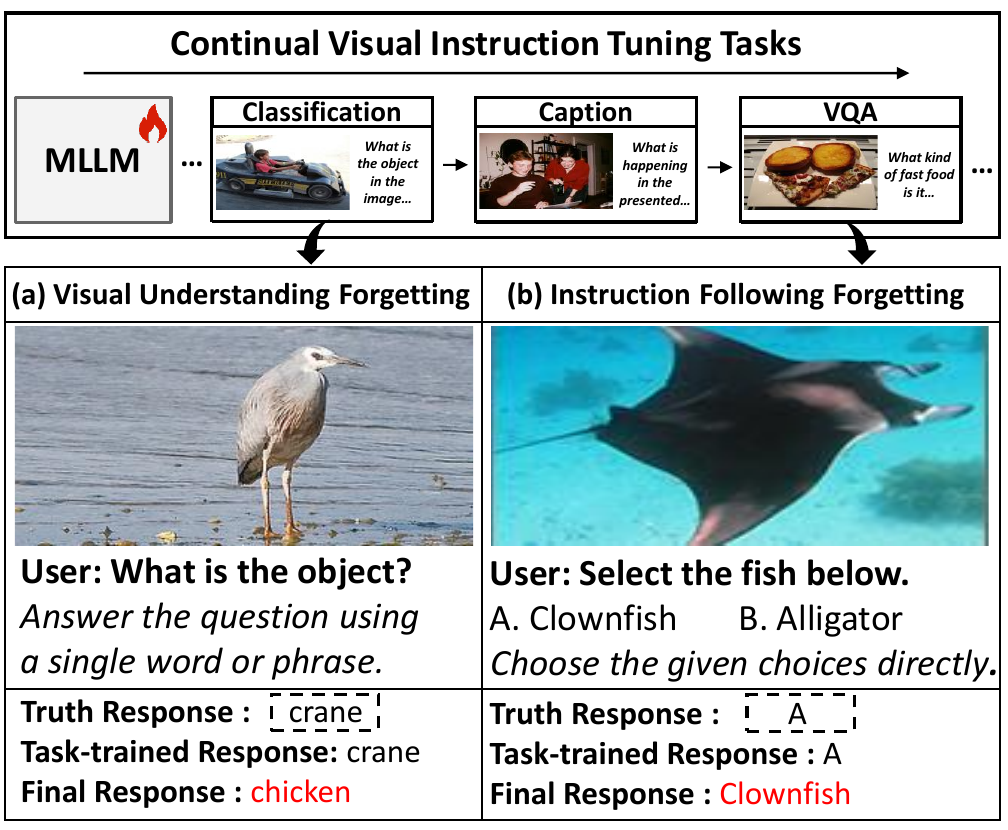}
   \caption{\textbf{Dual Catastrophic Forgetting.} In CVIT, both visual understanding ability (a) and instruction following ability (b) are subject to forgetting. \textit{Task-trained Response} refers to the output after fine-tuning on the specific task, while \textit{Final Response} refers to the output after fine-tuning on the last task.
   }
   \label{fig_1}
\end{figure}

Visual instruction tuning for MLLMs is typically performed within a static multitask learning framework \cite{dai2023instructblip, liu2024visual}, where a unified image-instruction-output data format is used. However, in real-world applications, MLLMs need to learn new vision-language tasks to accommodate evolving requirements. This practical necessity has sparked a growing interest in continual visual instruction tuning (CVIT), which allows MLLMs to incrementally learn new tasks, reducing the significant costs associated with retraining models from scratch. Despite its potential, existing studies \cite{zhai2023investigating, he2023continual} on continual learning have demonstrated that sequential fine-tuning can lead to catastrophic forgetting, where models forget previously learned knowledge when updated with new tasks. While recent work \cite{he2023continual, chen2024coin} has made strides in developing new benchmarks and approaches to mitigate forgetting, these efforts largely follow traditional continual learning paradigms and overlook the distinct challenges in  CVIT.

In this work, we identify a dual form of catastrophic forgetting in CVIT. First, the model often loses its previously acquired visual understanding when updated with new tasks ( See Fig.~\ref{fig_1} (a)). Second, its ability to follow instructions deteriorates as more tasks are learned over time (See Fig.~\ref{fig_1} (b)). Both forms of forgetting lead to performance degradation during the CVIT process. Inspired by the mixture-of-expert approaches \cite{shazeer2017outrageously, lin2024moe} and LoRA-based tuning approaches \cite{hu2021lora, liu2024dora}, we propose a new Separable Mixture-of-LoRA (SMoLoRA) approach, specifically designed to tackle the dual forgetting problem in CVIT.

In SMoLoRA, we introduce a new separable routing mechanism by leveraging two specialized modules: a visual understanding module and an instruction following module. This dual-domain approach enables specialized adaptation in each domain, allowing the model to preserve and refine both its visual and instruction-following abilities without interference. By dynamically selecting LoRA blocks tailored to the specific needs of each task, SMoLoRA effectively mitigates the dual forgetting problem. Additionally, we introduce a new benchmark to comprehensively evaluate CVIT performance of various approaches, overcoming the limitations of existing benchmarks \cite{he2023continual, chen2024coin}, such as restricted task and instruction diversity, insufficient evaluation on unseen datasets, and the lack of comprehensive metrics.

Our contributions are summarized as follows:
\begin{itemize}
\setlength{\itemsep}{0pt}
\setlength{\parsep}{0pt}
\setlength{\parskip}{0pt}

\item We uncover the issue of dual catastrophic forgetting in CVIT and introduce the SMoLoRA framework as a solution. Through a new separable routing mechanism, SMoLoRA dynamically selects 
task-specific LoRA blocks, optimizing both visual understanding and instruction-following capabilities, thus mitigating dual forgetting.

\item We propose a new CVIT benchmark that not only evaluates task-specific performance but also emphasizes generalization to unseen tasks and the model’s ability to handle diverse instructions, providing a more holistic assessment.

\item Through extensive experiments on our CVIT benchmark, we demonstrate that existing approaches struggle with dual forgetting and generalization to new tasks. In contrast, SMoLoRA excels across a wide range of metrics and scenarios, delivering superior performance.

\end{itemize}

%% file: sec_our/2_related.tex
\section{Related Work}
\label{sec:related}

\textbf{Visual instruction tuning.}
Recent advancements in Instruction Tuning \cite{ouyang2022training, zhang2023instruction} have significantly enhanced the ability of language models to better understand and accurately follow complex human instructions across a variety of contexts. Building upon this paradigm, Visual Instruction Tuning \cite{shen2024multimodal, he2024multi} further extends the capabilities by integrating both visual and textual data, enabling MLLMs to execute instructions that involve multiple data modalities. Models such as InstructBLIP \cite{dai2023instructblip}, LLaVA \cite{liu2024visual}, and MiniGPT-4 \cite{zhu2023minigpt} exemplify this approach by leveraging large-scale pre-training and sophisticated alignment techniques to unify vision and language understanding. SVIT \cite{zhao2023svit} constructs a large-scale dataset to enrich the diversity and informativeness of instruction-tuning data. Similarly, Vision-Flan \cite{xu2024vision} has developed a human-annotated instruction tuning dataset encompassing a broad spectrum of tasks, thereby enabling its application in advanced MLLMs. Unlike previous works that focus on static visual instruction tuning, this work addresses continual visual instruction tuning scenarios, where models need to adapt to new tasks over time without forgetting previous knowledge.

\textbf{Mixture of Experts.}
Mixture of Experts (MoE) is a successful deep learning approach \cite{jacobs1991adaptive, shazeer2017outrageously, lepikhin2020gshard, fedus2022switch} that increases model capacity and efficiency by dividing the model into specialized experts, activated by a gating network for specific inputs. In recent years, some works, \eg, \cite{shen2023scaling,lin2024moe,gou2023mixture}, have applied MoE to MLLMs to handle more complex multimodal tasks. VL-MoE \cite{shen2023scaling} uses sparsely-gated MoE to create specialized sub-models for task solving.  MoE-LLAVA \cite{lin2024moe} proposes an efficient training strategy called MoE-Tuning for LLAVA. MoCLE \cite{gou2023mixture} combines MoE with LoRA experts and a unique universal expert to activate task-specific model parameters, depending on clusters of instructions. Although MoE has garnered widespread attention in MLLMs, its performance in CVIT scenarios is underexplored. In this paper, we introduce SMoLoRA, a new MoE-based approach specifically designed for CVIT scenarios, and demonstrate its effectiveness on a CVIT benchmark.

\begin{figure*}[ht]
  \centering
   \includegraphics[width=\linewidth]{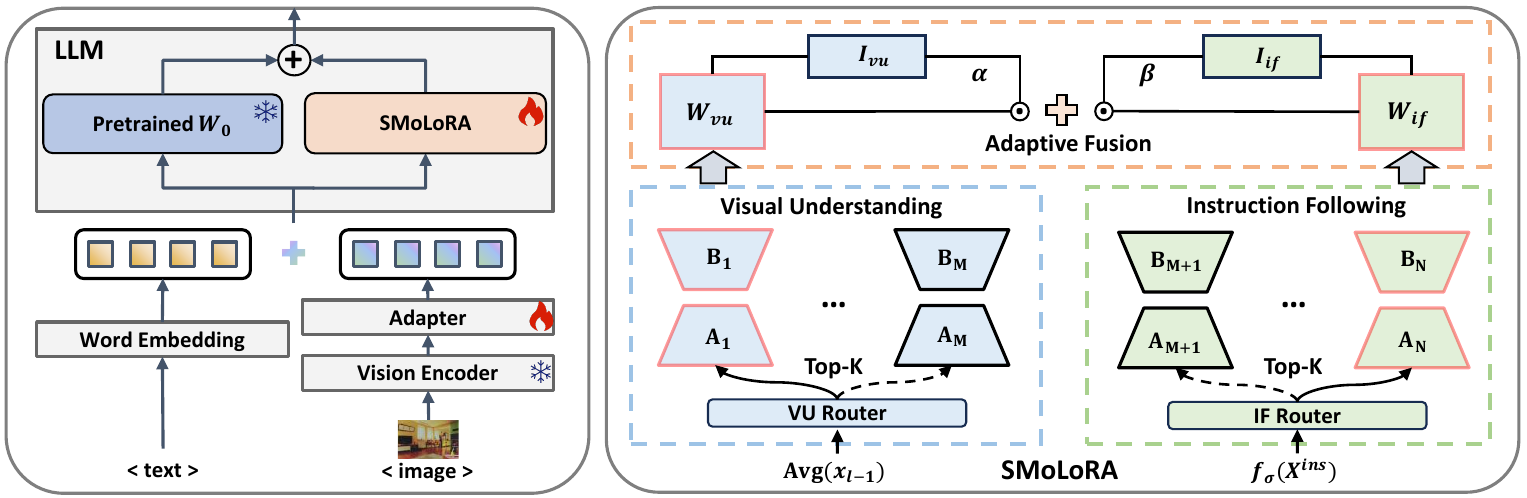}
   \caption{\textbf{SMoLoRA} is an effective method designed for CVIT, which can replace the original LoRA module in existing MLLMs. SMoLoRA consists of three modules. The \textbf{visual understanding} and \textbf{instruction following} modules use a separable routing strategy to select the most suitable LoRA blocks for the current input, thus preventing interference between dissimilar tasks. The \textbf{adaptive fusion} module performs weighted fusion on the results after separable routing using trainable parameters. Through the collaboration of these modules, SMoLoRA effectively addresses the problem of dual catastrophic forgetting in CVIT.
   }
   \label{fig_main}
\end{figure*}

\textbf{Continual learning for MLLMs.}
Continual Learning \cite{wang2024comprehensive, kirkpatrick2017overcoming, chaudhry2019tiny} enables models to incrementally acquire knowledge from new data while retaining previously learned information, thereby aligning with the requirement for MLLMs to stay synchronized with rapidly evolving human knowledge. EMT \cite{zhai2023investigating} first introduced the continual learning paradigm to MLLMs and identified catastrophic forgetting in classification tasks within these models. Eproj \cite{he2023continual} combined instruction tuning of MLLMs with continual learning and proposed a structural expansion approach to mitigate forgetting. Fwd-Prompt \cite{zheng2024beyond} addressed negative forward transfer during continual instruction tuning using a prompt-based solution. CoIN \cite{chen2024coin} applied the Mixture-of-Experts \cite{liu2023moelora, dou2023loramoe} approach to continual instruction tuning. However, while these methods have initiated continual learning in MLLMs, they do not fully address the diverse forms of catastrophic forgetting that arise during the process. In this work, we highlight the dual catastrophic forgetting challenge and introduce SMoLoRA, a new approach designed to effectively address this issue.

%% file: sec_our/3_method.tex
\section{Method}
\label{sec:method}

\subsection{Continual Visual Instruction Tuning}
\label{sec:3.1}
Given a pre-trained MLLM $f$, continual visual instruction tuning (CVIT) aims to incrementally fine-tune $f$ to learn a set of vision tasks $\mathrm{T} = \{\tau_1, \tau_2, \ldots, \tau_n\}$. At each time step $t$, the model is presented with a new task $\tau_t$ alongside a new dataset $\mathcal{D}_t$, to integrate this new knowledge into its existing capabilities. Typically, the dataset $\mathcal{D}_t$ follows the format $\{X^{ins}, X^{vis}, X^{ans}\}$, where $X^{ins}, X^{vis}, X^{ans}$ denote textual instruction input, visual input, and linguistic answer, respectively. However, when learning the new task $\tau_t$, previously seen data $\{\mathcal{D}_i\}_{i=1}^{t-1}$ is inaccessible, which introduces the challenge of dual catastrophic forgetting: 1) The model tends to forget previously learned visual understanding ability when updated with new data, and 2) Its ability to follow instructions gradually declines as additional tasks are learned. Next, we introduce an SMoLoRA approach to effectively mitigate the dual catastrophic forgetting problem in CVIT.

\subsection{Separable Mixture-of-LoRA}
\label{sec:3.2}

The MLLM $f$ typically consists of a vision encoder, an LLM, and an adapter. The vision encoder extracts visual features, which the adapter converts into word embeddings, enabling the LLM to interpret visual inputs (as shown in Fig.~\ref{fig_main}). During visual instruction tuning, only the LLM and adapter are updated, while the vision encoder remains frozen. For efficient task-specific updates within CVIT, we use LoRA fine-tuning. However, applying a single, shared LoRA across all tasks could lead to catastrophic forgetting in continual learning. To overcome this challenge, we propose a Separable Mixture-of-LoRA (SMoLoRA) approach tailored for CVIT, building upon the Mixture-of-LoRA (MoLoRA) framework \cite{liu2023moelora}.

MoLoRA, initially proposed for multitask learning, consists of a set of LoRA blocks along with a router network $G$. This router network dynamically selects different LoRA blocks based on the input, effectively allowing the model to specialize in different tasks. For a single LoRA, a low-rank update is applied by decomposing the adaptation matrix $\Delta W = BA$, where $B \in \mathbb{R}^{k \times r}$ and $A \in \mathbb{R}^{r \times d}$, with $r \ll \min(d, k)$. $d$ and $k$ are the input and output dimensions of the original weight matrix $W_0$ of the pre-trained model. For MoLoRA, the adaptation matrix $\Delta W$ is typically represented as:
\begin{equation}
\Delta W = \sum_{i=1}^{N} G(z)_i \cdot \Delta W_i,
\label{formula_molora}
\end{equation}
where $G(z)_i$ is the weight assigned to the $i$-th LoRA block predicted by the router network $G$ with the router input $z$. $\Delta W_i$ is the adaptation matrix of the $i$-th LoRA block. 

Extending MoLoRA from multitask learning to continual learning presents a complex dynamic routing challenge. In multitask learning, tasks often share common expert components based on task similarity. However, in continual learning, the model is exposed to an evolving stream of tasks with highly diverse and sometimes conflicting data distributions. This variability makes the task of selecting the right experts more nuanced and non-trivial. As new tasks emerge, the model must adapt its routing strategy to not only accommodate new data but also ensure that previously learned tasks are not negatively impacted. Otherwise, new tasks can overwrite or diminish the knowledge associated with previous ones, resulting in dual forgetting problem in the context of CVIT.

To address this challenge, SMoLoRA introduces the concept of separable routing, implemented through two distinct modules: a visual understanding module and an instruction following module, defined by:  
\begin{equation}
\Delta W^{vu} = \sum_{i=1}^{M} G^{vu}(z^{vu})_{i} \cdot \Delta W_i,
\label{formula_set_1}
\end{equation}
\begin{equation}
\Delta W^{if} = \sum_{j=M+1}^{N} G^{if}(z^{if})_{j} \cdot \Delta W_j.
\label{formula_set_2}
\end{equation}
where $z^{vu}$ and $z^{if}$ denote the router inputs of these two modules, respectively. $M$ represents the number of LoRA blocks for visual understanding module. The final output is generated by adaptively combining the outputs from these two adaptation matrices, allowing the model to effectively update itself for the new task. Suppose the input to the current linear layer is $x_{l-1}$, we can obtain the output of the layer $x_l$ through these adaptation matrices:
\begin{equation}
x_l =  W_0 x_{l-1}+\mathcal{F}(\Delta W^{vu}x_{l-1}, \Delta W^{if}x_{l-1})
\label{formula_set_3}
\end{equation} 
where $\mathcal{F}$ denotes the adaptive fusion approach. 

Unlike MoLoRA, which relies on input tokens to infer task-specific information, our routing approach leverages both visual understanding and instruction following to more precisely identify the type of input instance. This dual-domain analysis allows for specialized adaptation in each domain, ensuring that the model can maintain both capabilities without interference. By dynamically selecting experts tailored to the task's requirements, our method effectively addresses the dual forgetting problem. Next, we elaborate on separable routing and adaptive fusion. 

\subsection{Separable Routing and Adaptive Fusion}
\label{sec:3.3}
\textbf{Separable routing.} In visual understanding, the model analyzes both the visual features of the image and the key information guided by the text. Therefore, we employ an instance-based routing method that prioritizes the overall information from the instance inputs. For the current linear layer input $x_{l-1} \in \mathbb{R}^{d \times s}$, we compute the average feature of the current instance, denoted as $\text{Avg}(x_{l-1})$, across the sequence dimension $s$. We maintain a router matrix $R^{vu} \in \mathbb{R}^{M \times d}$ and let $z^{vu}= \text{Avg}(x_{l-1})$ to select LoRA blocks for the input to the visual understanding module:
\begin{equation}
G^{vu}(z^{vu}) = \underset{\text{dim}=0}{\operatorname{softmax}}\left( \operatorname{top}_k\left( R^{vu}\text{Avg}(x_{l-1})   \right) \right),
\label{formula_wgk}
\end{equation}
where \( \operatorname{top}_k(\cdot) \) selects the top \( k \) values, setting the rest to \( -\infty \), and \( \operatorname{softmax}(\cdot) \) normalizes these values. 

In contrast, the differences in instruction requirements also play a crucial role in contributing to task discrepancies. The instruction following module leverages the embedding of the current task instruction, extracted using Sentence-BERT \cite{reimers2019sentence}, \( f_{\sigma}(X^{ins}) \in \mathbb{R}^{e \times 1} \), as the input \( z^{if} \) for routing. This approach helps mitigate interference between different task requirements during continual learning:
\begin{equation}
G^{if}(z^{if}) = \underset{\text{dim}=0}{\operatorname{softmax}}\left( \operatorname{top}_k\left( R^{if} f_{\sigma}(X^{ins})   \right) \right),
\label{formula_wif}
\end{equation}
where \( R^{if} \in \mathbb{R}^{(N-M) \times e} \) is the router matrix maintained by the instruction following module. 

\begin{figure*}[ht]
  \centering
   \includegraphics[width=\linewidth]{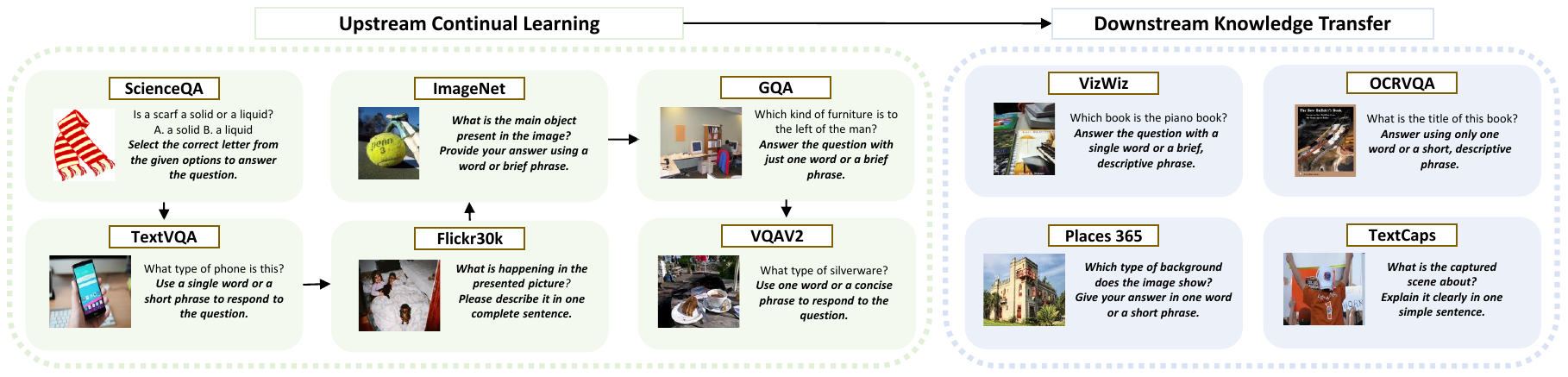}
   \caption{\textbf{Our designed CVIT benchmark.} It consists of two parts: upstream tasks assess model stability during continual learning, while downstream tasks evaluate transfer generalization to unseen tasks. We incorporate diverse instructions across tasks and design corresponding metrics to evaluate the model's performance on this benchmark.
   }
   \label{fig_bench}
   \vspace{-5mm}
\end{figure*}

\textbf{Adaptive fusion.} Considering that the contributions of the two modules' router strategies may not be identical during learning, we adopt an adaptive approach to assign weights to the outputs of these two modules. 
First, we compute the outputs of the two modules $x_{l}^{vu}$ and $x_{l}^{if} \in \mathbb{R}^{k \times s}$ separately: 
\begin{equation}
x_{l}^{vu} = \Delta W^{vu}x_{l-1}, \quad x_{l}^{if} = \Delta W^{if}x_{l-1}
\label{formula_two_out}
\end{equation}

Let \( I^{vu} \in \mathbb{R}^{1 \times k} \) and \( I^{if} \in \mathbb{R}^{1 \times k} \) represent the trainable importance matrices of the visual understanding module and the instruction following module, respectively. The weight ratio  \( [\alpha, \beta] \) can be expressed as:
\begin{equation}
[\alpha, \beta]^T = \underset{\text{dim}=-1} {\operatorname{softmax}}\left( \operatorname{concat}\left( I^{vu} x_{l}^{vu} , I^{if} x_{l}^{if}  \right) \right),
\label{eq:weight_ratio}
\end{equation}
where ${\alpha, \beta \in \mathbb{R}^{1 \times s}}$, \( \operatorname{concat}(\cdot) \) represents concatenation along the first dimension. The final result is given by:
\begin{equation}
\mathcal{F}(x_{l}^{vu}, x_{l}^{if}) = \alpha \circ x_{l}^{vu} + \beta \circ x_{l}^{if},
\label{formula_fuse}
\end{equation}
where $\circ$ represents the element-wise multiplication applied independently to each row of the matrices.

%% file: sec_our/4_benchmark.tex
\section{CVIT Benchmark}
\label{sec:benchmark}

\subsection{Benchmark Construction}
\label{sec:4.1}
While previous studies \cite{he2023continual, chen2024coin} have established benchmarks for Continual Visual Instruction Tuning (CVIT) in MLLMs, they have yet to fully address key challenges, such as generalizing to unseen tasks, adapting to diverse instructions, and assessing instruction following ability. To address these challenges, we present our CVIT benchmark, as shown in Fig.~\ref{fig_bench}. Our benchmark construction can be described from three key aspects:

\textbf{Dataset collection and standardization.}
Our CVIT benchmark integrates widely used vision-language datasets, covering diverse task types such as VQA, image classification, and image captioning. In response to the issue of varying input-output formats across different task types (e.g., the lack of language guidance in classification tasks), we standardize all tasks into a unified textual instruction format to accommodate a sequential stream of tasks, thus ensuring a consistent transformation into a text-to-text format \cite{zhang2023citb}.

\textbf{Continual stability and transfer generalization.}
The CVIT benchmark is used in a two-stage manner: upstream continual learning and downstream knowledge transfer. For the upstream tasks, we sequentially select ScienceQA \cite{lu2022learn}, TextVQA \cite{singh2019towards}, Flickr30k \cite{plummer2015flickr30k}, ImageNet \cite{deng2009imagenet}, GQA \cite{hudson2019gqa}, and VQAv2 \cite{goyal2017making} as subjects for continual learning and stability evaluation. Regarding the downstream tasks, we evaluate the zero- and few-shot performance across four diverse datasets after completing all upstream tasks. VizWiz \cite{gurari2018vizwiz}, TextCaps \cite{sidorov2020textcaps}, and OCRVQA \cite{mishra2019ocr} are evaluated under zero-shot settings, whereas Places365 \cite{zhou2017places} is evaluated under both 5-shot and 10-shot per-class settings.

\textbf{Exploration of instruction diversity.}
To more thoroughly assess model adaptability and robustness, we incorporate both single-type and multi-type instructions into each task, which mirrors real-world scenarios and allows for a systematic analysis of the impact of instruction diversity on model performance. It also enables evaluating various CVIT methods' adaptability under conditions involving both instruction types. More details about CVIT benchmark are provided in Section D of the supplementary material.

\subsection{Metrics Designing}
\label{sec:4.2}
To evaluate the CVIT benchmark, we utilize multiple metrics to assess overall performance and the degree of forgetting. Let $a_{k,j}$ denote the accuracy on the $j$-th task (where $j < k$) after fine-tuning on the $k$-th task. We define the performance at the current stage and across all previous stages as Average Performance (AP) and Mean Average Performance (MAP), respectively, following training on $k$ tasks:
\begin{equation}
\mathrm{AP}_{k}=\frac{1}{k} \sum_{j=1}^{k} {a}_{k, j}, \quad
\mathrm{MAP}_{k}=\frac{1}{k} \sum_{i=1}^{k} \mathrm{AP}_{i}.
\label{formula_AP}
\end{equation}

In assessing the model's degree of forgetting, backward transfer \cite{wang2024comprehensive} is employed as a metric:
\begin{equation}
\mathrm{BWT}_{k}=\frac{1}{k-1} \sum_{j=1}^{k-1} ({a}_{k, j}-{a}_{j,j}).
\label{formula_BWT}
\end{equation}

\begin{table*}[ht]
\caption{The evaluated results (\%) on upstream continual learning for our CVIT benchmark after tuning on the final task. *: We follow the settings of the CoIN \cite{chen2024coin} and randomly select 100 classes from ImageNet as experimental data. $^\dagger$: The performance of DirLoRA is typically considered to be an upper bound for CVIT \cite{zheng2024beyond}. Our method achieves the best results across all settings.}
\label{results_main_IF}
\aboverulesep=0pt
\belowrulesep=0pt
\renewcommand\arraystretch{1.3}
\renewcommand\tabcolsep{4.0pt}
\centering
\resizebox{\textwidth}{!}{
\begin{tabular}{cc|cccccc|cccc}
\toprule[1.2pt]
\multirow{2}{*}{} &
\multirow{2}{*}{\textbf{Method}} & 
\multicolumn{6}{c}{\textbf{Accuracy on Each Task}} &
\multicolumn{4}{c}{\textbf{Overall Results}} \\ 
{} &{}  & ScienceQA & TextVQA & Flickr30k & ImageNet* & GQA & VQAv2 & AP $\uparrow$ & MAP $\uparrow$ & BWT $\uparrow$ & MIF $\uparrow$\\ 
\midrule[1pt]
\multirow{12}{*}{\textbf{Single-type}}
& {Multitask } & 83.49 & 61.93 & 169.21 & 96.53 & 60.07 & 65.80 & 89.51 & - & -  & 98.38 \\ 
& {Zero-shot } & 52.72 & 2.95 & 52.64 & 22.10 & 2.73 & 0.65 & 22.30 & - & -  & 17.84 \\  
&  {DirLoRA$^\dagger$ } &  83.75 &  60.66 &  164.20 &  96.71 &  58.55 &  64.93 &  88.13 &  - &  0.00  &  98.41 \\ \cline{2-12}
& {SeqLoRA \cite{hu2021lora}} & 55.31 & 50.22 & 33.89 & 22.73 & 50.52 & 64.61 & 46.21 & 57.41 & -48.10  & 78.35 \\ 
& {MoeLoRA \cite{chen2024coin}} & 55.01 & 48.87 & 32.04 & 22.00 & 50.03 & 63.64 & 45.27 & 56.16 & -48.05  & 79.97 \\

& {DoRA \cite{liu2024dora}} & 51.26 & 46.36 & 36.41 & 28.24 & 45.29 & 56.87 & 44.07 & 65.03 & -31.12  & 78.59 \\ 
& {C-LoRA \cite{smith2023continual}} & 57.25  & 38.70  & 56.50  & 25.27  & 42.89  & 54.06  &45.78  & 57.04  & -19.58  & 65.84  \\
& {Replay \cite{chaudhry2019tiny}} & 75.61 & 47.58 & 31.97 & 35.84 & 48.51 & 58.67 & 49.70 & 69.78 & -22.71  & 82.06 \\ 
& {EWC \cite{kirkpatrick2017overcoming}} & 57.04 & 50.02 & 32.96 & 22.85 & 50.16 & 64.54 & 46.26 & 56.19 & -49.71  & 78.90 \\ 
& {EWC+TIR \cite{he2023continual}} & 72.22  & 44.78  & 34.54  &25.98  & 46.86  & 58.73  & 47.19  & 67.21  & -25.64  & 81.62 	 \\ 

& {Eproj \cite{he2023continual}} & 65.29  & 52.87  & 148.19  & 39.45  & 28.06  & 57.86  &65.29  &73.53  & -14.02  & 89.81 	 \\ \cline{2-12}
& \cellcolor[gray]{0.9} \textbf{SMoLoRA(Ours)} & \cellcolor[gray]{0.9} \textbf{77.36} & \cellcolor[gray]{0.9} \textbf{58.29} & \cellcolor[gray]{0.9} \textbf{151.99} & \cellcolor[gray]{0.9} \textbf{95.35} & \cellcolor[gray]{0.9} \textbf{51.96} & \cellcolor[gray]{0.9} \textbf{65.71} & \cellcolor[gray]{0.9} \textbf{83.44} & \cellcolor[gray]{0.9} \textbf{84.85} & \cellcolor[gray]{0.9} \textbf{-3.23}  &  \cellcolor[gray]{0.9} \textbf{97.79} \\ 
\midrule[1pt] \midrule[1pt]
\multirow{12}{*}{\textbf{Multi-type}}
& {Multitask } & 82.83 & 61.27 & 175.97 & 96.44 & 59.67 & 65.80 & 90.33 & - & - & 97.93 \\ 
& {Zero-shot } & 51.85 & 5.11 & 44.05 & 20.34 & 2.37 & 1.16 & 20.81 & - & - & 19.45 \\ 
&   {DirLoRA$^\dagger$ } &  83.85 &  60.51 &  164.66 &  96.71 &  57.93 &  64.90 &  88.09 &  - &  0.00  &  98.31 \\ \cline{2-12}
& {SeqLoRA \cite{hu2021lora}} & 59.21 & 50.80 & 20.99 & 20.30 & 49.98 & 64.41 & 44.28 & 53.75 & -48.73 & 79.47 \\ 
& {MoeLoRA \cite{chen2024coin}} & 58.09 & 53.30 & 22.82 & 22.61 & 51.80 & 65.15 & 45.63 & 54.88 & -49.94 & 78.19 \\
& {DoRA} \cite{liu2024dora} & 52.03 & 47.37 & 27.97 & 26.18 & 46.05 & 57.33 & 42.82 & 56.24 & -34.11 & 78.30 \\ 
& {C-LoRA \cite{smith2023continual}} & 55.58  & 38.64  &59.05  &22.81  & 40.93  & 51.65  & 44.78  & 52.37  & -18.83  & 70.01  \\ 

& {Replay \cite{chaudhry2019tiny}} & 66.06 & 47.78 & 24.21 & 25.66 & 46.53 & 58.59 & 44.81 & 66.68 & -26.88 & 80.38 \\ 
& {EWC \cite{kirkpatrick2017overcoming}} & 53.60 & 49.07 & 20.38 & 20.48 & 50.11 & 64.63 & 43.10 & 53.94 & -52.47  & 78.18 \\ 
& {EWC+TIR \cite{he2023continual}} & 66.94  & 45.76  &29.49  & 21.68  & 46.90  &58.80  & 44.93  & 64.51  & -26.38  & 80.56 	 	 \\ 

& {Eproj \cite{he2023continual}} & 63.45  & 53.18  & 151.41  & 20.63  & 45.30  & 57.32  & 65.22  & 72.10  & -14.43  & 89.93  \\ \cline{2-12} \cline{2-12}
& \cellcolor[gray]{0.9} \textbf{SMoLoRA(Ours)} & \cellcolor[gray]{0.9} \cellcolor[gray]{0.9} \textbf{80.50} & \cellcolor[gray]{0.9} \textbf{58.30} & \cellcolor[gray]{0.9} \textbf{146.63} & \cellcolor[gray]{0.9} \textbf{94.28} & \cellcolor[gray]{0.9} \textbf{52.42} & \cellcolor[gray]{0.9} \textbf{65.96} & \cellcolor[gray]{0.9} \textbf{83.02} & \cellcolor[gray]{0.9} \textbf{85.05} & \cellcolor[gray]{0.9} \textbf{-6.50} & \cellcolor[gray]{0.9} \textbf{98.12} \\ \bottomrule[1.2pt]

\end{tabular}
}
\end{table*}

Additionally, to further evaluate the model's instruction following ability after fine-tuning the $k$-th task, we design a new Mean Instruction Following (MIF) metric:
\begin{equation}
\mathrm{MIF}_{k}=\frac{1}{k} \sum_{j=1}^{k} \left(\frac{1}{n}\sum_{i=1}^{n} \mathcal{B}_j(o_i^j)\right),
\label{formula_MIF}
\end{equation}
where $\mathcal{B}()$ is a binary decision function that outputs 1 if the result  $o_i^j$ matches the current task's instruction format, and 0 otherwise. $n$ denotes the number of test samples.

%% file: sec_our/5_exp.tex
\section{Experiment}
\label{sec:experiment}

\subsection{Experiment Setup}
\textbf{Comparison methods.}
We compare our method against a diverse set of alternatives to highlight its superior performance. \textbf{Multitask} and \textbf{Zero-Shot} represent training all tasks jointly and no task-specific training, respectively. \textbf{DirLoRA} introduces a separate LoRA for each individual task, while \textbf{SeqLoRA} \cite{hu2021lora} sequentially fine-tunes all tasks using a single shared LoRA. We also include the enhanced LoRA variants, \textbf{DoRA} \cite{liu2024dora} and \textbf{C-LoRA} \cite{smith2023continual}, for comparison. Additionally, we evaluate classic continual learning approaches such as \textbf{EWC} \cite{kirkpatrick2017overcoming} and \textbf{Replay} \cite{chaudhry2019tiny}. Finally, we assess a series of methods specifically designed for CVIT, including \textbf{EWC+TIR} \cite{he2023continual}, \textbf{Eproj} \cite{he2023continual}, and a token-wise MoLoRA method, \textbf{MoeLoRA} \cite{chen2024coin}.

\textbf{Implemented details.}
We use the pre-trained first-stage \textbf{LLaVA-v1.5-7B} \cite{liu2024visual} as our base model. The hyperparameters for SMoLoRA are configured as follows: a learning rate of $1 \times 10^{-4}$ with a cosine decay schedule, and a batch size of 64. Training is performed over a single epoch. The number of LoRA blocks is set to 4 for both visual understanding module and instruction following module, with each LoRA having a rank of 16, and \( \operatorname{top}_k(\cdot) \) set to \( \operatorname{top}_1(\cdot) \). Both SMoLoRA and the baseline methods are applied specifically to the feedforward network (FFN) layers of the LLM and the adapter. We also conduct relevant experiments on MiniGPT-4 \cite{zhu2023minigpt}, as detailed in supplementary material C.

\begin{figure*}[ht]
  \centering
   \includegraphics[width=\linewidth]{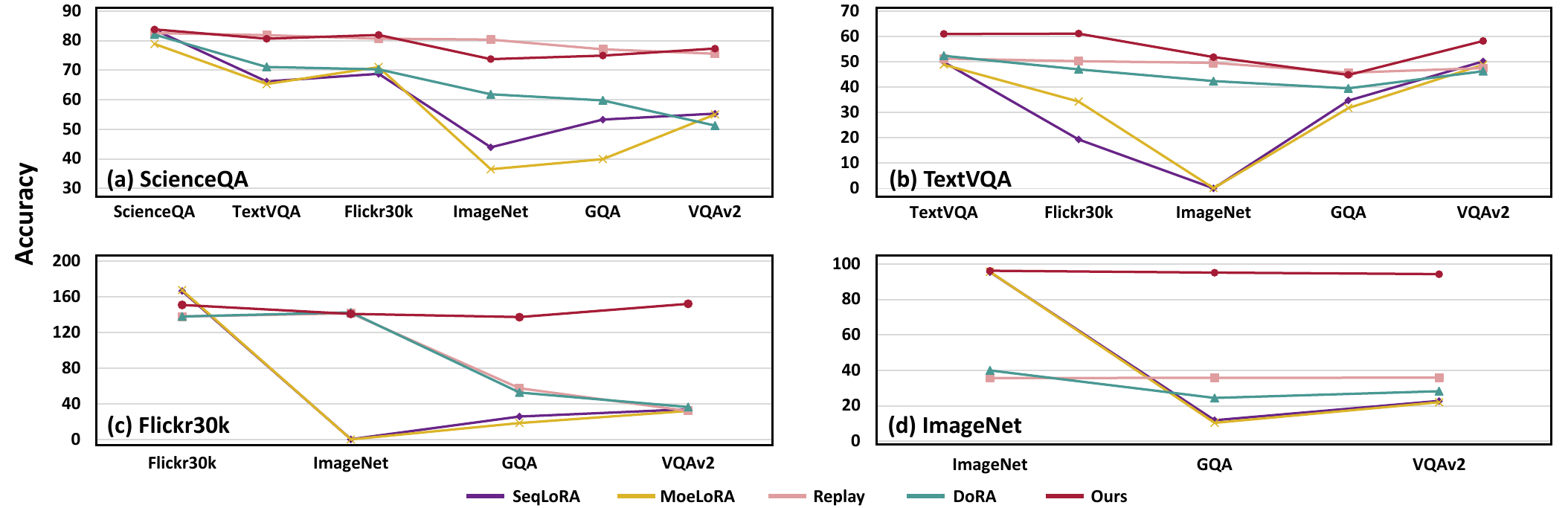}
   \caption{Accuracy (\%) variations across the first four datasets at different stages of our CVIT benchmark. Our method consistently sustains high performance during continual learning.
   }
   \label{fig_f4}
\end{figure*}

\begin{table}[t]
\caption{The zero-shot results (\%) on downstream knowledge transfer for our CVIT benchmark. Our method outperforms other methods across datasets, metrics, and instruction types, showcasing its stronger zero-shot generalization ability.}
\label{results_down_zero}
\aboverulesep=0pt
\belowrulesep=0pt
\renewcommand\arraystretch{1.3}
\renewcommand\tabcolsep{4.0pt}
\centering
\resizebox{\columnwidth}{!}{
\begin{tabular}{cc|ccc|cc}
\toprule[1.2pt]
\multirow{2}{*}{} &
\multirow{2}{*}{\textbf{Method}} &
\multicolumn{3}{c|}{\textbf{Zero-shot}} &
\multicolumn{2}{c}{\textbf{Overall Results}} \\ 
{}  &{} & VizWiz & TextCaps & OCRVQA & AP & MIF \\ \midrule[1pt]
\multirow{5}{*}{\textbf{\makecell{Single- \\ type}}}
& {SeqLoRA \cite{hu2021lora}} & 27.69 & 9.73 & 30.97 & 22.80 & 61.58 \\ 
& {MoeLoRA \cite{chen2024coin}} & 29.64 & 8.74 & 30.27 & 22.88 & 60.11 \\
& {Replay \cite{chaudhry2019tiny}} & 29.54 & 7.52 & 37.78 & 24.95 & 61.10 \\ 
& {DoRA \cite{liu2024dora}} & 29.03 & 9.45 & 37.14 & 25.21 & 61.98 \\ \cline{2-7}
& \cellcolor[gray]{0.9} \textbf{SMoLoRA(Ours)} & \cellcolor[gray]{0.9} \textbf{32.72} & \cellcolor[gray]{0.9} \textbf{30.97} & \cellcolor[gray]{0.9} \textbf{39.37}  & \cellcolor[gray]{0.9} \textbf{34.35} & \cellcolor[gray]{0.9} \textbf{90.94} \\ 
\midrule[1pt] \midrule[1pt]

\multirow{5}{*}{\textbf{\makecell{Multi- \\ type}}}
& {SeqLoRA \cite{hu2021lora}} & 22.27 & 7.10 & 31.06  & 20.14 & 61.09 \\ 
& {MoeLoRA \cite{chen2024coin}} & 26.30 & 7.13 & 32.55 & 21.99 & 61.28 \\
& {Replay \cite{chaudhry2019tiny}} & 24.70 & 8.32 & 38.18  & 23.73 & 61.33 \\ 
& {DoRA \cite{liu2024dora}} & 26.16 & 8.83 & 38.08 & 24.36 & 61.59 \\ \cline{2-7}
& \cellcolor[gray]{0.9} \textbf{SMoLoRA(Ours)} & \cellcolor[gray]{0.9} \textbf{27.46} & \cellcolor[gray]{0.9} \textbf{25.28} & \cellcolor[gray]{0.9} \textbf{39.86} & \cellcolor[gray]{0.9} \textbf{30.87} & \cellcolor[gray]{0.9} \textbf{85.76} \\  
\bottomrule[1.2pt]

\end{tabular}
}
\end{table}

\subsection{Main Results}
\textbf{Upstream tasks.}
In the upstream tasks, we perform continual fine-tuning on the model sequentially, starting from ScienceQA and progressing to VQAv2, under both single- and multi-type instruction settings. We evaluate model performance after fine-tuning on VQAv2. As shown in Table~\ref{results_main_IF}, our method consistently outperforms alternative approaches by a substantial margin, achieving superior results on both individual tasks and overall evaluation metrics.

In the single-type instruction setting, each dataset follows a single instruction format. Compared to MoeLoRA, our approach achieves impressive improvements in terms of task-solving ability: +38.17\% in AP, +28.69\% in MAP, and +17.82\% in MIF. Regarding the BWT metric, while other methods show a forgetting range between -40\% and -20\%, our method maintains a minimal BWT of just -3.23\%. Notably, our method also outperforms traditional continual learning approaches like EWC and Replay across all metrics. In the multi-type instruction setting, the increased complexity of instructions poses additional challenges for learning. However, our method consistently demonstrates superior performance over the baselines.

Fig.~\ref{fig_f4} further illustrates the performance trends across the first four datasets throughout the continual instruction tuning process. It is clear that our method consistently sustains high performance during continual learning, while other approaches experience abrupt performance drops due to catastrophic forgetting. See supplementary material B for performance variation details.

\textbf{Downstream tasks.}
We conduct experiments on four downstream unseen tasks to evaluate the generalization capabilities of our method in knowledge transfer. Table~\ref{results_down_zero} presents a comparison of our method with other approaches under the zero-shot setting, across both single-type instruction and multi-type instruction. In the single-type setting, our method outperforms the second-best approach on three unseen datasets, with improvements of 3.08\%, 21.24\%, and 1.59\%, respectively, while achieving a notable 90.94\% on the MIF metric. In the few-shot setting, as shown in Table~\ref{results_down_few}, our method further boosts performance after fine-tuning on a small sample. It surpasses the second-best approach by 10.90\% and 9.30\% on the AP metric for single-type and multi-type settings, respectively. These results highlight that our continual tuning approach significantly enhances the model's ability to generalize to unseen tasks.

\subsection{Ablation Study}
\textbf{Effect of the components in SMoLoRA.}
We begin by examining the impact of each component in SMoLoRA on the single-type CVIT benchmark. The VU, IF, and AF modules represent the visual understanding, instruction following, and adaptive fusion components, respectively. Without any of these modules is equivalent to SeqLoRA. In contrast, when the VU or IF module is added, performance improves significantly, with the IF module yielding the most substantial gains, particularly in the MIF metric. This underscores the critical role of the instruction following in task learning within CVIT. When both the VU and IF modules are combined by averaging their outputs, performance improves further. The inclusion of the AF module provides an additional boost, emphasizing its effectiveness in enhancing the model's overall capabilities. We further show how the AF module balances the proportional relationship between the VU and IF modules in supplementary material A.

\begin{table}[t]
\caption{The 5/10-shot results (\%) on downstream knowledge transfer for our CVIT benchmark. Our method achieves the best results across metrics and instruction types, showcasing its stronger few-shot generalization ability.}
\label{results_down_few}
\aboverulesep=0pt
\aboverulesep=0pt
\belowrulesep=0pt
\renewcommand\arraystretch{1.3}
\renewcommand\tabcolsep{4.0pt}
\centering
\small
\resizebox{0.8\columnwidth}{!}{
\begin{tabular}{cc|cc|cc}
\toprule[1.2pt]
\multirow{2}{*}{} &
\multirow{2}{*}{\textbf{Method}} &
{\textbf{5-shot}} &
{\textbf{10-shot}} &
\multicolumn{2}{c}{\textbf{Overall Results}} \\ 
{} &{} & \multicolumn{2}{c|}{Places365}  & AP & MIF \\ 
\midrule[1pt]

\multirow{5}{*}{\textbf{\makecell{Single- \\ type}}} 
& {SeqLoRA \cite{hu2021lora}} & 13.15 & 12.34 & 12.75 & 99.90 \\ 
& {MoeLoRA \cite{chen2024coin}} & 18.58 & 21.38 & 19.98 & 99.82 \\
& {Replay \cite{chaudhry2019tiny}} & 17.80 & 21.02 & 19.41 & 99.87 \\ 
& {DoRA \cite{liu2024dora}} & 28.30 & 30.53 & 29.42 & 100.00 \\ \cline{2-6}
& \cellcolor[gray]{0.9} \textbf{SMoLoRA(Ours)} & \cellcolor[gray]{0.9} \textbf{36.95} & \cellcolor[gray]{0.9} \textbf{43.68}  & \cellcolor[gray]{0.9} \textbf{40.32} & \cellcolor[gray]{0.9} \textbf{100.00} \\ 
\midrule[1pt] \midrule[1pt]

\multirow{5}{*}{\textbf{\makecell{Multi- \\ type}}} 
& {SeqLoRA \cite{hu2021lora}} & 17.82 & 17.98  & 17.9 & 99.90 \\ 
& {MoeLoRA \cite{chen2024coin}} & 10.56 & 11.12 & 10.84 & 2.98 \\
& {Replay \cite{chaudhry2019tiny}} & 17.38 & 22.34  & 19.86 & 99.92 \\ 
& {DoRA \cite{liu2024dora}} & 29.55 & 32.24 & 30.90 & 100.00 \\ \cline{2-6}
& \cellcolor[gray]{0.9} \textbf{SMoLoRA(Ours)} & \cellcolor[gray]{0.9} \textbf{37.00} & \cellcolor[gray]{0.9} \textbf{43.40} & \cellcolor[gray]{0.9} \textbf{40.20} & \cellcolor[gray]{0.9} \textbf{100.00} \\ \bottomrule[1.2pt]

\end{tabular}}
\vspace{-2mm}
\end{table}

\begin{table}[t]
\small
  \centering
  \caption{Ablation studies of different modules on the single-type CVIT benchmark. The \textbf{VU}, \textbf{IF}, and \textbf{AF} modules represent the visual understanding, instruction following, and adaptive fusion components, respectively.}
    {
    \begin{tabular}{ccc|ccccc}
    \toprule[1.0pt]
   
   VU &IF &AF     
   &{AP} &{MAP} &{BWT} &{MIF}\\ \midrule[0.8pt]
  $\times$ &$\times$ &$\times$   &46.21  &57.41 &-48.10 &78.35 \\
  $\checkmark$ &$\times$ &$\times$  &53.49  &67.84 &-33.06 &80.12 \\
  $\times$ &$\checkmark$ &$\times$  &71.97  &79.56 &-17.42 &\textbf{98.38}\\
  $\checkmark$ &$\checkmark$ &$\times$  &75.16  &78.72 &-10.99 &97.43 \\
  \rowcolor[gray]{0.9} $\checkmark$ &$\checkmark$ &$\checkmark$  &\textbf{83.44}  &\textbf{84.85} &\textbf{-3.23} &{97.79}\\ \bottomrule[1.0pt]
    \end{tabular}}
  \label{tab_ablation}

\end{table}

\begin{figure}[t]
  \centering
   \includegraphics[width=\linewidth]{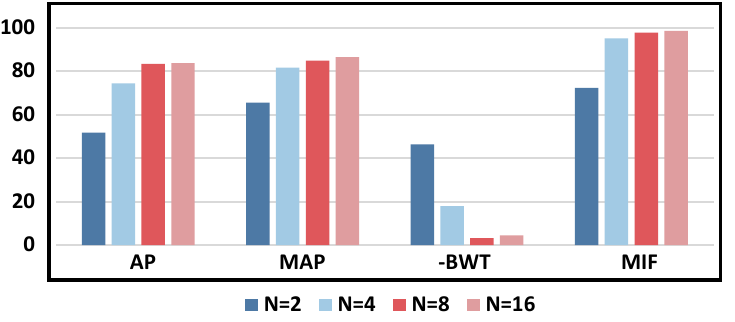}
   \caption{Evaluation results under varying numbers of LoRA blocks settings. As the number of LoRA blocks increases, the model's performance improves across all metrics.
   }
   \label{fig_number_unit}
   \vspace{-6mm}
\end{figure}

\textbf{Effect of varying the number of LoRA blocks.}
Fig.~\ref{fig_number_unit} illustrates the impact of varying the number of LoRA blocks in SMoLoRA. As the number of LoRA blocks increases, the model's performance improves across all metrics. This improvement occurs because a greater number of modules provides more options for task-specific selection, reducing interference from dissimilar tasks. However, when the number of LoRA blocks reaches higher values (e.g., 8 or 16), performance stabilizes, indicating that additional LoRA blocks do not significantly enhance learning. This suggests that further increasing the number of modules beyond a certain point may lead to inefficient resource usage without providing substantial benefits.

\begin{figure}[t]
  \centering
   \includegraphics[width=\linewidth]{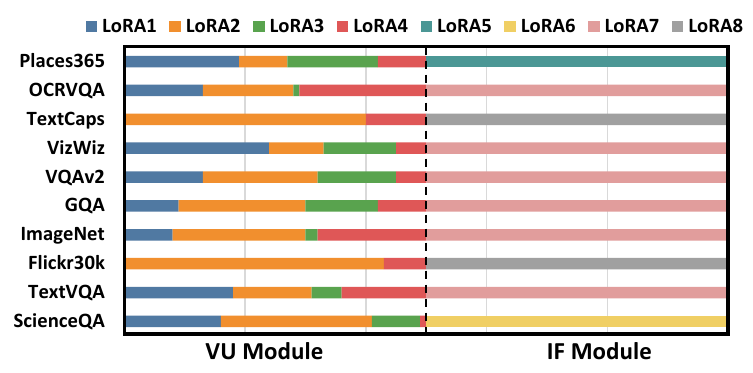}
   \caption{Routing behavior across the VU module and IF module for different datasets. The visualization results show the reliable routing of our SMoLoRA method across different tasks.
   }
   \label{fig_flu_select}
   \vspace{-2mm}
\end{figure}

\textbf{Effect of routing of LoRA blocks.}
To further explore how SMoLoRA mitigates the dual catastrophic forgetting issue, we randomly select a fixed number of examples from each dataset and visualize their LoRA blocks routing behaviors in Fig.~\ref{fig_flu_select}. Longer color bars indicate a stronger preference for specific LoRA blocks. Within the VU module’s four LoRA blocks, we observe considerable variation in selections across different tasks. However, similar tasks, such as the captioning tasks of Flickr30k and TextCaps, show converging routing patterns. In the IF module, where routing is driven by task-specific instructions, tasks with differing instructional requirements demonstrate distinct preferences for different LoRA blocks. This visualization highlights the precise control our method exerts during the routing process, which effectively mitigates forgetting by preventing improper LoRA block allocations across tasks. 

\vspace{-0.5mm}

\begin{figure}[t]
  \centering
   \includegraphics[width=\linewidth]{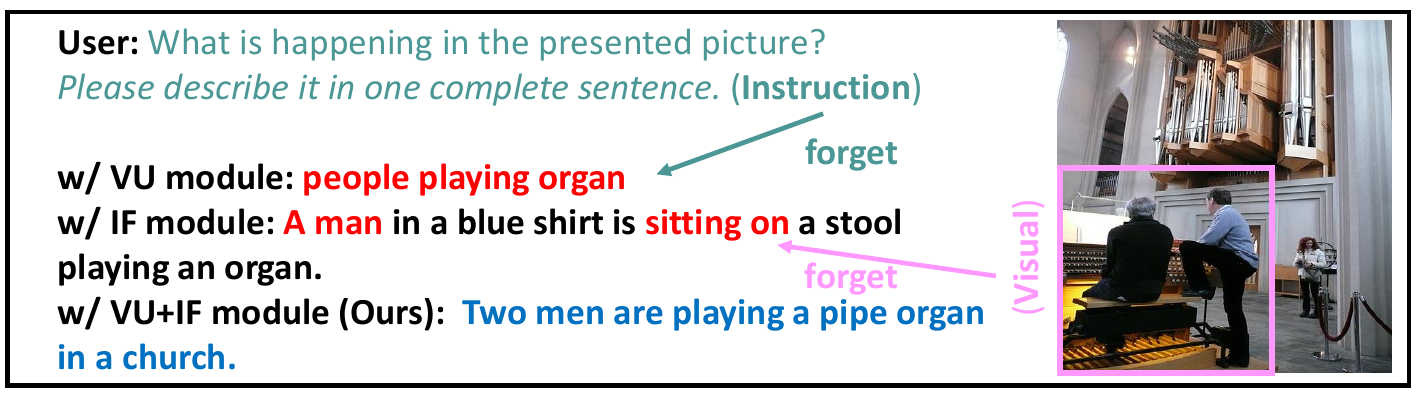}
   \caption{\textbf{Case study}. \textcolor{red}{Red} text indicates inappropriate outputs, while \textcolor{blue_1}{blue} text represents appropriate outputs.
   }
   \label{fig_case}
   \vspace{-6mm}
\end{figure}

\subsection{Case Study}
Finally, we present a case study in Fig.~\ref{fig_case} to illustrate how SMoLoRA effectively addresses dual forgetting issues through the integration of two specialized modules: the Visual Understanding (VU) module, designed to mitigate visual understanding forgetting, and the Instruction Following (IF) module, aimed at tackling instruction following forgetting. When only the VU module is applied, the model comprehends visuals but fails to generate coherent sentences, producing fragmented words instead. Conversely, with only the IF module, the model adheres to the required text format but misinterprets visual targets. Integrating both modules achieves robust performance for both visual understanding and instruction adherence. Additional case studies can be found in Section E of the supplementary material.

%% file: sec_our/6_conclusion.tex
\section{Conclusion}
\label{sec:conclusion}
In this paper, we identify a dual form of catastrophic forgetting in CVIT and propose an effective SMoLoRA method. SMoLoRA leverages separable routing to dynamically assign appropriate LoRA blocks to different tasks, thereby preserving the model's visual understanding and instruction following capabilities, mitigating the dual catastrophic forgetting. Additionally, we introduce a new CVIT benchmark that evaluates not only model accuracy and forgetting on standard tasks but also its ability to generalize to unseen tasks and adapt to diverse instructions. Extensive experiments on the CVIT benchmark demonstrate the superior performance of our SMoLoRA over existing methods.

%% file: sec_our/X_suppl.tex
\clearpage
\setcounter{page}{1}
\maketitlesupplementary

\section*{A. Effect of Weight Ratio}
To further investigate the impact of the weight ratio $[\alpha,\beta]$ within the adaptive fusion module, we randomly sampled the values of $\alpha$ and $\beta$ during inference across different layers of the fine-tuned model, as illustrated in Fig.~\ref{fig_weight_ratio}. The analysis reveals that $\alpha$ and $\beta$ exhibit significant variations across layers. Furthermore, we computed the average values of $\alpha$ and $\beta$ for all layers. The resulting ratios indicate that the instruction following module plays a more pivotal role in the routing process compared to the visual understanding module.

\section*{B. Results at Different Stages}
As shown in Table~\ref{details_Seq}-\ref{details_SMo}, we present the experimental results of SeqLoRA and our SMoLoRA method at various stages from ScienceQA to VQAv2 on the CVIT benchmark. The results demonstrate that with each additional training stage, our approach consistently maintains stable and superior performance across different datasets.

\section*{C. More Experiments on MLLMs }
To further assess the versatility of our SMoLoRA method, we implemented it on another advanced MLLM, MiniGPT-4 \cite{zhu2023minigpt}, as shown in Table~\ref{results_minigpt}. While MiniGPT-4 exhibits a lower degree of forgetting compared to LLAVA, our method still yields significant overall performance enhancements compared to SeqLoRA.  Notably, on the MIF metric, we observed improvements of 17.98\% and 20.90\% in single- and multi-type instruction settings, respectively. These results underscore the effectiveness of our method in mitigating forgetting in instruction following.

\section*{D. Details of CVIT Benchmark} \label{sup_D}
Table~\ref{datasets} provides an overview of the CVIT benchmark, detailing the instruction templates for each dataset and the number of samples.

\textbf{ScienceQA \cite{lu2022learn}: }ScienceQA is a comprehensive dataset consisting of science-related questions and answers designed to evaluate and enhance the reasoning and problem-solving capabilities of AI models in scientific domains.

\textbf{TextVQA \cite{singh2019towards}: }TextVQA is a visual question answering dataset that focuses on questions requiring models to read and understand text embedded within images to provide accurate answers.

\textbf{Flickr30k \cite{plummer2015flickr30k}: } Flickr30k is a large-scale image dataset containing over 30,000 photos sourced from Flickr, each annotated with multiple descriptive captions, widely used for training and evaluating image captioning and vision-language models.

\textbf{ImageNet \cite{deng2009imagenet}: }ImageNet is a large-scale visual dataset containing millions of annotated images across thousands of object categories, widely used for training and evaluating computer vision models.

\textbf{GQA \cite{hudson2019gqa}: }GQA is a visual question-response dataset comprising complex compositional questions about images, designed to evaluate and enhance AI models' reasoning and relational understanding capabilities in interpreting visual content.

\textbf{VQAv2 \cite{goyal2017making}: }VQAv2 is a widely-used visual question answering dataset that consists of images paired with diverse questions and multiple corresponding answers, designed to assess and enhance the ability of models to understand and reason about visual information.

\textbf{VizWiz \cite{gurari2018vizwiz}: }VizWiz originates from real-world visual question answering scenarios in which visually impaired individuals capture images and pose spoken questions about them. Each visual question is accompanied by 10 crowdsourced answers, facilitating the development of assistive technologies for the visually impaired community.

\textbf{TextCaps \cite{sidorov2020textcaps}: }TextCaps is an image captioning dataset that requires models to generate descriptive captions by reading and interpreting text embedded within images, thereby enhancing the ability to incorporate textual information into visual descriptions.

\textbf{OCRVQA \cite{mishra2019ocr}: }OCRVQA is a visual question answering dataset designed to evaluate the ability of the models to read and comprehend text embedded within images to accurately answer related questions.

\textbf{Places365 \cite{zhou2017places}: }Places365 is a large-scale scene recognition dataset comprising over 1.8 million images across 365 diverse scene categories, widely used for training and evaluating computer vision models in understanding and classifying various environments.

\section*{E. Additional Case Studies} \label{sup_e}
As shown in Fig.~\ref{fig_weight_ratio}, we present additional cases to validate the efficacy of our proposed SMoLoRA. Individual modules (VU and IF) are only capable of resolving their respective forgetting problems, whereas the combination of both can simultaneously mitigate the dual catastrophic forgetting.


\begin{figure*}[ht]
  \centering
   \includegraphics[width=\linewidth]{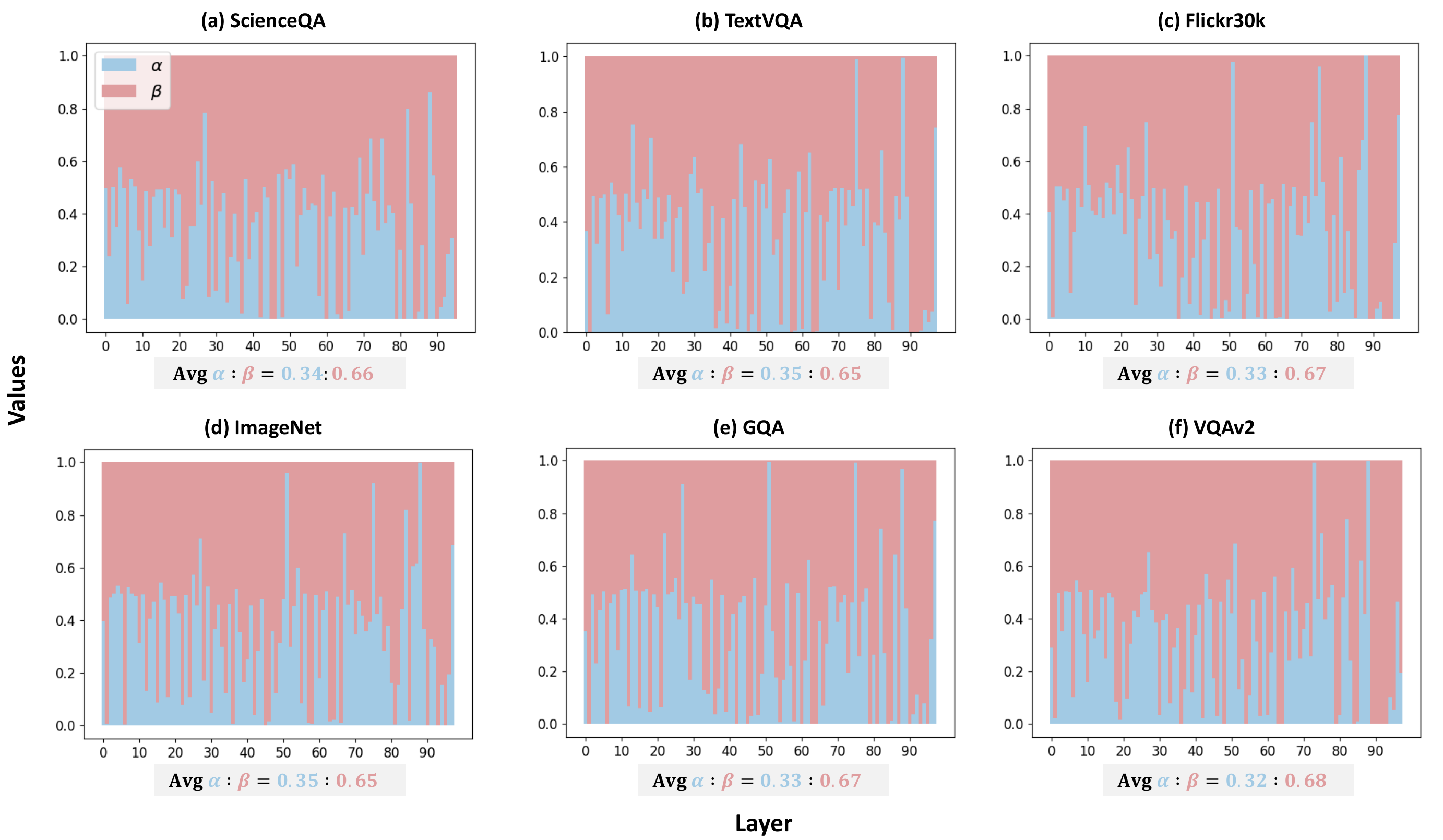}
   \caption{Distribution of weight ratio $[\alpha,\beta]$ across different layers.
The visualization results show that the instruction following module ($\beta$) plays a more critical role in the routing process than the visual understanding module ($\alpha$).
   }
   \label{fig_weight_ratio}
\end{figure*}

\begin{table}[t]
\caption{The evaluated results (\%) of \textbf{SeqLoRA} for upstream continual learning \textbf{at different stages} of our CVIT benchmark. Each row represents a different training stage.}
\label{details_Seq}
\renewcommand\arraystretch{1.3}
\renewcommand\tabcolsep{4.0pt}
\resizebox{\columnwidth}{!}{
\begin{tabular}{ccccccc}
\toprule[1.2pt]
 & ScienceQA & TextVQA & Flickr30k & ImageNet & GQA & VQAv2  \\ \midrule[1pt]
\multirow{6}{*}{\textbf{\makecell{Single- \\ type}}} & 83.75 &  &  &  &  &    \\ \cline{2-7}
 & 66.21 & 49.95 &  &  &  &    \\ \cline{2-7}
 & 68.78 & 19.35 & 166.33  &  &  &    \\ \cline{2-7}
 & 43.90 & 0.05 & 0.26  & 95.45 &  &    \\ \cline{2-7}
 & 53.29  & 34.74 & 25.76 & 11.84 & 57.69  &   \\ \cline{2-7}
 & 55.31 & 50.22 & 33.89  & 22.73 & 50.52  & 64.37  \\ \midrule[1pt] \midrule[1pt]

 \multirow{6}{*}{\textbf{\makecell{Multi- \\ type}}} & 83.85 &  &  &  &  &    \\ \cline{2-7}
 & 69.18 & 50.24 &  &  &  &    \\  \cline{2-7}
 & 51.38 & 0.00 & 156.85  &  &  &    \\ \cline{2-7}
 & 37.77  & 0.01 & 0.13  & 95.98 &  &    \\ \cline{2-7}
 & 53.01  & 33.78 & 3.77 & 10.18 & 58.01  &   \\ \cline{2-7}
 & 59.21 & 50.80 & 20.99  & 20.30 & 49.98 & 64.41  \\ 
 \bottomrule[1.2pt]

 \end{tabular}}
\end{table}

\begin{table}[t]
\caption{The evaluated results (\%) of our \textbf{SmoLoRA} for upstream continual learning \textbf{at different stages} of our CVIT benchmark. Each row represents a different training stage.}
\label{details_SMo}
\renewcommand\arraystretch{1.3}
\renewcommand\tabcolsep{4.0pt}
\resizebox{\columnwidth}{!}{
\begin{tabular}{ccccccc}
\toprule[1.2pt]
 & ScienceQA & TextVQA & Flickr30k & ImageNet & GQA & VQAv2  \\ \midrule[1pt]
\multirow{6}{*}{\textbf{\makecell{Single- \\ type}}} & 83.85 &  &  &  &  &    \\ \cline{2-7}
 & 80.71 & 61.05 &  &  &  &    \\ \cline{2-7}
 & 81.99 & 61.20 & 150.72  &  &  &    \\ \cline{2-7}
 & 73.80  & 51.90 & 140.71  & 96.28 &  &    \\ \cline{2-7}
 & 74.98  & 44.87 & 137.08 & 95.45 & 59.19  &   \\ \cline{2-7}
 & 77.36 & 58.29 & 151.99  & 95.35 & 51.96  & 65.71  \\ \midrule[1pt] \midrule[1pt]

 \multirow{6}{*}{\textbf{\makecell{Multi- \\ type}}} & 84.53 &  &  &  &  &    \\ \cline{2-7}
 & 80.71 & 61.24 &  &  &  &    \\  \cline{2-7}
 & 81.58 & 60.24 & 162.78  &  &  &    \\ \cline{2-7}
 & 74.44  & 44.28 & 136.92  & 96.14 &  &    \\ \cline{2-7}
 & 78.09  & 45.31 & 133.03 & 95.09 & 59.96  &   \\ \cline{2-7}
 & 80.50 & 58.30 & 146.63  & 94.28 & 52.42 & 65.96  \\ 
 \bottomrule[1.2pt]

 \end{tabular}}
\end{table}

\begin{table*}[ht]
\caption{The evaluated results (\%) on upstream continual learning for our CVIT benchmark using \textbf{MiniGPT-4} after tuning on the final task.}
\label{results_minigpt}
\aboverulesep=0pt
\belowrulesep=0pt
\renewcommand\arraystretch{1.3}
\renewcommand\tabcolsep{4.0pt}
\centering
\resizebox{\textwidth}{!}{
\begin{tabular}{cc|cccccc|cccc}
\toprule[1.2pt]
\multirow{2}{*}{} &
\multirow{2}{*}{\textbf{Method}} & 
\multicolumn{6}{c}{\textbf{Accuracy on Each Task}} &
\multicolumn{4}{c}{\textbf{Overall Results}} \\ 
{} &{}  & ScienceQA & TextVQA & Flickr30k & ImageNet* & GQA & VQAv2 & AP $\uparrow$ & MAP $\uparrow$ & BWT $\uparrow$ & MIF $\uparrow$\\ 
\midrule[1pt]
\multirow{3}{*}{\textbf{Single-type}}
&  {Zero-shot } & 41.10 & 0.00 & 0.03 & 26.32 & 0.00 & 0.00 & 11.24 & - & -  & 0.96 \\  
& {SeqLoRA \cite{hu2021lora}} & \textbf{54.56} & 36.18 & 134.83 & 38.93 & 40.85 & 34.16 & 56.59 & 56.78 & \textbf{9.22}  & 58.42 \\

& \cellcolor[gray]{0.9} \textbf{SMoLoRA(Ours)} & \cellcolor[gray]{0.9} 54.16 & \cellcolor[gray]{0.9} \textbf{38.61} & \cellcolor[gray]{0.9} \textbf{135.60} & \cellcolor[gray]{0.9} \textbf{48.46} & \cellcolor[gray]{0.9} \textbf{44.11} & \cellcolor[gray]{0.9} \textbf{48.35} & \cellcolor[gray]{0.9} \textbf{61.55} & \cellcolor[gray]{0.9} \textbf{62.05} & \cellcolor[gray]{0.9} 3.39 &  \cellcolor[gray]{0.9} \textbf{76.40} \\ 
\midrule[1pt] \midrule[1pt]
\multirow{3}{*}{\textbf{Multi-type}}
&  {Zero-shot } &  42.72 & 0.00 & 0.01 & 26.83 & 0.00 & 0.00 & 11.59 & - & -  & 1.18 \\  
& {SeqLoRA \cite{hu2021lora}} & 54.61 & 34.24 & \textbf{116.09} &30.32 & 40.65 & 28.91 & 50.80 & 51.76 & -2.37  & 54.49 \\ 
& \cellcolor[gray]{0.9} \textbf{SMoLoRA(Ours)} & \cellcolor[gray]{0.9} \textbf{57.72} & \cellcolor[gray]{0.9} \textbf{39.60} & \cellcolor[gray]{0.9} 112.43 & \cellcolor[gray]{0.9} \textbf{43.27} & \cellcolor[gray]{0.9} \textbf{44.65} & \cellcolor[gray]{0.9} \textbf{38.43} & \cellcolor[gray]{0.9} \textbf{56.02} & \cellcolor[gray]{0.9} \textbf{60.67} & \cellcolor[gray]{0.9} \textbf{1.84}  &  \cellcolor[gray]{0.9} \textbf{75.39} \\ \bottomrule[1.2pt]

\end{tabular}
}
\end{table*}

\begin{table*}[ht]
\caption{Details of datasets in our CVIT Benchmark.}
\label{datasets}
\centering
\resizebox{\textwidth}{!}{
\begin{tabular}{|c|c|c|c|c}
\hline
\textbf{Dataset} & \textbf{Instruction template} & \textbf{Train Number} & \textbf{Test Number} \\ \hline \hline
\multirow{5}{*}{\textbf{ScienceQA}} & \textless{}question\textgreater{}\textless{}instruction\textgreater{}: Answer with the option's letter from the given choices directly. &
\multirow{5}{*}{\textbf{12726}} & \multirow{5}{*}{\textbf{4241}} \\
 & \textless{}question\textgreater{}\textless{}instruction\textgreater{}: Select the correct answer by choosing the corresponding letter from the options provided. & & \\
 & \textless{}question\textgreater{}\textless{}instruction\textgreater{}: Select the correct letter from the given options to answer the question. & & \\
 & \textless{}question\textgreater{}\textless{}instruction\textgreater{}: Identify the correct answer by choosing the appropriate letter from the choices. & & \\
 & \textless{}question\textgreater{}\textless{}instruction\textgreater{}: Pick the correct answer by selecting the letter associated with the correct choice. & & \\ \hline
\multirow{5}{*}{\textbf{TextVQA}} & \textless{}question\textgreater{}\textless{}instruction\textgreater{}: Answer using only one word or a short, descriptive phrase. & \multirow{5}{*}{\textbf{34602}} & \multirow{5}{*}{\textbf{5000}} \\
 & \textless{}question\textgreater{}\textless{}instruction\textgreater{}: Use a single word or a short phrase to respond to the question. & & \\
 & \textless{}question\textgreater{}\textless{}instruction\textgreater{}: Use one word or a concise phrase to respond to the question. & & \\
 & \textless{}question\textgreater{}\textless{}instruction\textgreater{}: Answer the question with just one word or a brief phrase. & & \\
& \textless{}question\textgreater{}\textless{}instruction\textgreater{}:Answer the question with a single word or a brief, descriptive phrase. & & \\\hline
\multirow{5}{*}{\textbf{Flickr30k}}& \textless{}instruction\textgreater{}: What is depicted in the displayed picture? Summarize it using a single, concise sentence. & \multirow{5}{*}{\textbf{145000}} & \multirow{5}{*}{\textbf{1014}} \\
 & \textless{}instruction\textgreater{}: What is happening in the presented picture? Please describe it in one complete sentence. & & \\
 & \textless{}instruction\textgreater{}: What does the image display clearly and succinctly? Provide a full sentence explaining it. & & \\
 & \textless{}instruction\textgreater{}: How would you interpret the scene in the picture? Express your answer in one informative sentence. & & \\
 & \textless{}instruction\textgreater{}: What is the captured scene about? Explain it clearly in one simple sentence. & & \\ \hline
\multirow{5}{*}{\textbf{ImageNet}} & \textless{}instruction\textgreater{}: What is the main object present in the image? Provide your answer using a word or brief phrase. &
\multirow{5}{*}{\textbf{117715}} & \multirow{5}{*}{\textbf{5050}} \\
 & \textless{}instruction\textgreater{}: Which specific object does the image depict? Give your answer in one word or a short phrase. & & \\
 & \textless{}instruction\textgreater{}: What category does the object in the image belong to? Answer using a single word or phrase. & & \\
 & \textless{}instruction\textgreater{}: What is the object in the image? Answer briefly with a word or a short phrase. & & \\
 & \textless{}instruction\textgreater{}: What is the primary object visible in the image? Answer briefly with a word or a short phrase. & & \\ \hline
\multirow{5}{*}{\textbf{GQA}} & \textless{}question\textgreater{}\textless{}instruction\textgreater{}: Answer using only one word or a short, descriptive phrase. &
\multirow{5}{*}{\textbf{72140}} & \multirow{5}{*}{\textbf{12578}}\\
 & \textless{}question\textgreater{}\textless{}instruction\textgreater{}: Use a single word or a short phrase to respond to the question. & & \\
 & \textless{}question\textgreater{}\textless{}instruction\textgreater{}: Use one word or a concise phrase to respond to the question. & & \\
& \textless{}question\textgreater{}\textless{}instruction\textgreater{}: Answer the question with just one word or a brief phrase. & & \\
& \textless{}question\textgreater{}\textless{}instruction\textgreater{}:Answer the question with a single word or a brief, descriptive phrase. & & \\\hline
\multirow{5}{*}{\textbf{VQAv2}} & \textless{}question\textgreater{}\textless{}instruction\textgreater{}: Answer using only one word or a short, descriptive phrase. &
\multirow{5}{*}{\textbf{82783}} & \multirow{5}{*}{\textbf{53588}}\\
 & \textless{}question\textgreater{}\textless{}instruction\textgreater{}: Use a single word or a short phrase to respond to the question. & & \\
 & \textless{}question\textgreater{}\textless{}instruction\textgreater{}: Use one word or a concise phrase to respond to the question. & & \\
& \textless{}question\textgreater{}\textless{}instruction\textgreater{}: Answer the question with just one word or a brief phrase. & & \\
& \textless{}question\textgreater{}\textless{}instruction\textgreater{}:Answer the question with a single word or a brief, descriptive phrase. & & \\\hline\hline

\multirow{5}{*}{\textbf{VizWiz}} & \textless{}question\textgreater{}\textless{}instruction\textgreater{}: Answer using only one word or a short, descriptive phrase. &
\multirow{5}{*}{\textbf{0}} & \multirow{5}{*}{\textbf{4319}}\\
 & \textless{}question\textgreater{}\textless{}instruction\textgreater{}: Use a single word or a short phrase to respond to the question. & & \\
 & \textless{}question\textgreater{}\textless{}instruction\textgreater{}: Use one word or a concise phrase to respond to the question. & & \\
& \textless{}question\textgreater{}\textless{}instruction\textgreater{}: Answer the question with just one word or a brief phrase. & & \\
& \textless{}question\textgreater{}\textless{}instruction\textgreater{}:Answer the question with a single word or a brief, descriptive phrase. & & \\ \hline
\multirow{5}{*}{\textbf{TextCaps}}& \textless{}instruction\textgreater{}: What is depicted in the displayed picture? Summarize it using a single, concise sentence. & \multirow{5}{*}{\textbf{0}} & \multirow{5}{*}{\textbf{15830}} \\
 & \textless{}instruction\textgreater{}: What is happening in the presented picture? Please describe it in one complete sentence. & & \\
 & \textless{}instruction\textgreater{}: What does the image display clearly and succinctly? Provide a full sentence explaining it. & & \\
 & \textless{}instruction\textgreater{}: How would you interpret the scene in the picture? Express your answer in one informative sentence. & & \\
 & \textless{}instruction\textgreater{}: What is the captured scene about? Explain it clearly in one simple sentence. & & \\ \hline

\multirow{5}{*}{\textbf{OCRVQA}} & \textless{}question\textgreater{}\textless{}instruction\textgreater{}: Answer using only one word or a short, descriptive phrase. &
\multirow{5}{*}{\textbf{0}} & \multirow{5}{*}{\textbf{99926}}\\
 & \textless{}question\textgreater{}\textless{}instruction\textgreater{}: Use a single word or a short phrase to respond to the question. & & \\
 & \textless{}question\textgreater{}\textless{}instruction\textgreater{}: Use one word or a concise phrase to respond to the question. & & \\
& \textless{}question\textgreater{}\textless{}instruction\textgreater{}: Answer the question with just one word or a brief phrase. & & \\
& \textless{}question\textgreater{}\textless{}instruction\textgreater{}:Answer the question with a single word or a brief, descriptive phrase. & & \\\hline
\multirow{5}{*}{\textbf{Places365}} & \textless{}instruction\textgreater{}: What is the background of the image? Answer the question using a single word or phrase. &
\multirow{5}{*}{\textbf{5 \& 10}} & \multirow{5}{*}{\textbf{36500}} \\
 & \textless{}instruction\textgreater{}: What is the background depicted in the image? Provide your answer using a word or brief phrase. & & \\
 & \textless{}instruction\textgreater{}: Which type of background does the image show? Give your answer in one word or a short phrase. & & \\
 & \textless{}instruction\textgreater{}: What category best describes the background in the image? Answer using a brief phrase or single word. & & \\
 & \textless{}instruction\textgreater{}: What is the primary background visible in the image? Answer briefly with a word or a short phrase. & & \\ \hline
\end{tabular}
}
\end{table*}

\begin{figure*}[ht]
  \centering
   \includegraphics[width=0.9\linewidth]{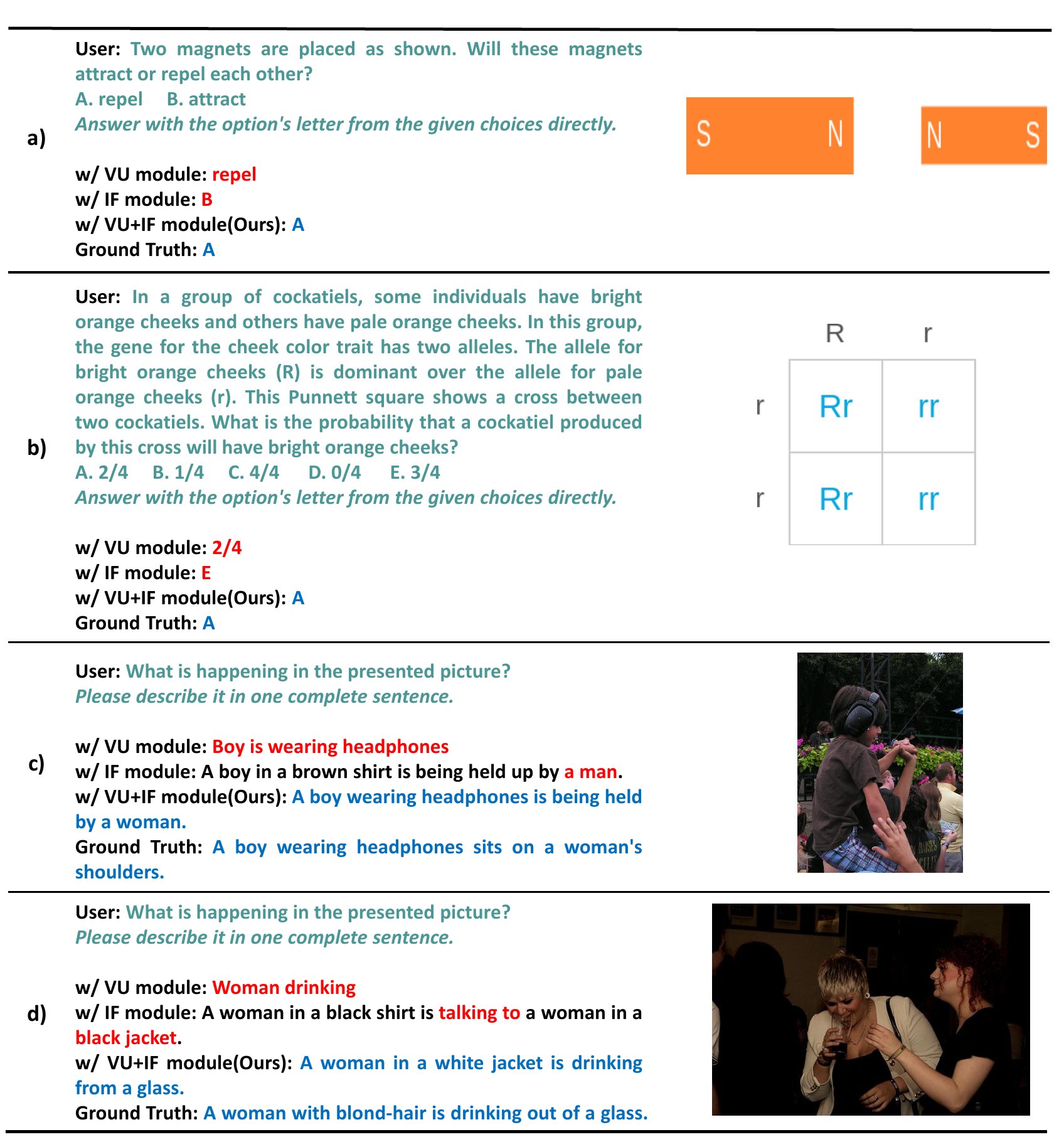}
   \caption{Additional Case Studies on the Effectiveness of Separation in SMoLoRA.}
   \label{fig_weight_ratio}
\end{figure*}